%% file: cvpr2024_Temporal3D.tex
\documentclass[final]{cvpr}

\usepackage{times}
\usepackage{epsfig}
\usepackage{graphicx}
\usepackage{amsmath}
\usepackage{amssymb}

\usepackage{bm}
\usepackage{multirow}
\usepackage{multicol}
\usepackage{floatrow}
\usepackage[normalem]{ulem}
\usepackage{graphics}
\usepackage{algpseudocode}
\usepackage{caption}
\usepackage{makecell}
\usepackage{booktabs}
\usepackage{scalerel}
\usepackage{paralist}
\usepackage[table]{xcolor}
\usepackage{color,colortbl}
\usepackage{amssymb}
\usepackage{pifont}
\newcommand{\cmark}{\ding{51}}%
\newcommand{\xmark}{\ding{55}}%
\usepackage[accsupp]{axessibility}

\usepackage[pagebackref=true,breaklinks=true,colorlinks,bookmarks=false,citecolor=blue,linkcolor=blue]{hyperref}

\usepackage{xcolor}
\definecolor{gray}{rgb}{0.5,0.5,0.5} 
\definecolor{frenchblue}{rgb}{0.0, 0.45, 0.73}
\definecolor{gray}{rgb}{0.5,0.5,0.5} 
\definecolor{green}{rgb}{0, 0.4, 0} 
\definecolor{orange}{rgb}{1, 0.5, 0} 	
\definecolor{mahogany}{rgb}{0.75, 0.25, 0.0}
\definecolor{purple}{rgb}{0.6, 0, 0.6}
\definecolor{darkgreen}{rgb}{0, 0.4, 0.4} 
\definecolor{teal}{rgb}{0.0, 0.5, 0.5}
\definecolor{aaaa}{rgb}{0.55, 0.1, 0.7}
\definecolor{red}{rgb}{1.0, 0, 0}
\definecolor{plotpurple}{rgb}{0.2353, 0.2, 0.90196}
\definecolor{plotorange}{rgb}{1.0, 0.6, 0.2}
\definecolor{plotgreen}{rgb}{0.2, 0.784313, 0.2}
\definecolor{plotred}{rgb}{1.0, 0.2, 0.392}
\definecolor{lightgray}{gray}{0.9}
\definecolor{LightCyan}{rgb}{0.88,1,1}
\definecolor{baselinecolor}{gray}{.9}

\newboolean{revising}
\setboolean{revising}{true}

\def\cvprPaperID{} 
\def\confName{CVPR}
\def\confYear{2024}
\graphicspath{{./figs/}{./figs/result/}}

\begin{document}

\title{PTT: Point-Trajectory Transformer for Efficient Temporal 3D Object Detection}

\author{Kuan-Chih Huang$^1$ \quad Weijie Lyu$^1$ \quad Ming-Hsuan Yang$^{1,2}$ \quad Yi-Hsuan Tsai$^2$ \vspace{0.2cm}\\
$^1$University of California, Merced \quad $^2$Google 
}

\maketitle

\urlstyle{same}

\pagestyle{empty}  
\thispagestyle{empty}

\begin{abstract}
    Recent temporal LiDAR-based 3D object detectors achieve promising performance based on the two-stage proposal-based approach. 
    They generate 3D box candidates from the first-stage dense detector, followed by different temporal aggregation methods.
    However, these approaches require per-frame objects or whole point clouds, posing challenges related to memory bank utilization.
    Moreover, point clouds and trajectory features are combined solely based on concatenation, which may neglect effective interactions between them.
    In this paper, we propose a point-trajectory transformer with long short-term memory for efficient temporal 3D object detection.
    To this end, we only utilize point clouds of current-frame objects and their historical trajectories as input to minimize the memory bank storage requirement.
    Furthermore, we introduce modules to encode trajectory features, focusing on long short-term and future-aware perspectives, and then effectively aggregate them with point cloud features.
    We conduct extensive experiments on the large-scale Waymo dataset to demonstrate that our approach performs well against state-of-the-art methods.
    %
    Code and models will be made publicly available at \url{https://github.com/kuanchihhuang/PTT}.
\end{abstract}

\input{sec/0_Introduction}

\input{sec/1_RelatedWork}

\input{sec/2_Approach}

\input{sec/3_Experiments}

\input{sec/4_Conclusion}

\vspace{-4mm}
\paragraph{Acknowledgements.}
This work was supported in part by the Intelligence Advanced Research Projects Activity (IARPA) via Department of Interior/ Interior Business Center (DOI/IBC) contract number 140D0423C0074. The U.S. Government is authorized to reproduce and distribute reprints for Governmental purposes notwithstanding any copyright annotation thereon. Disclaimer: The views and conclusions contained herein are those of the authors and should not be interpreted as necessarily representing the official policies or endorsements, either expressed or implied, of IARPA, DOI/IBC, or the U.S. Government.


{\small
\bibliographystyle{ieee_fullname}
\bibliography{egbib}
}

\end{document}

%% file: sec/0_Introduction.tex
\section{Introduction}

Detecting 3D objects~\cite{huang2022monodtr, ma2019am3d, shi2019pointrcnn, Lang2019pointpillars, he2020sassd, huang2023vgw3d} is a crucial problem for autonomous driving, enabling vehicles to perceive and interpret their surroundings accurately.
With the advent of LiDAR sensors, significant strides have been made in enhancing the precision of 3D object detection.
Nevertheless, the inherent sparsity of point cloud data captured by LiDAR sensors poses a significant challenge. The sparse data complicates accurately identifying and localizing objects, particularly in challenging occlusion scenarios.

Recent research studies~\cite{3d-man, mppnet, msf, Zhou_centerformer} have shifted their focus towards utilizing multi-frame LiDAR data as input to enhance detection performance. This approach leverages point cloud sequences captured over time as the vehicle moves in real-world scenarios. Compared to single-frame information, these sequences offer a more comprehensive and denser representation of the surrounding environment. 

\begin{figure}
    \centering

\includegraphics[width=1\linewidth]{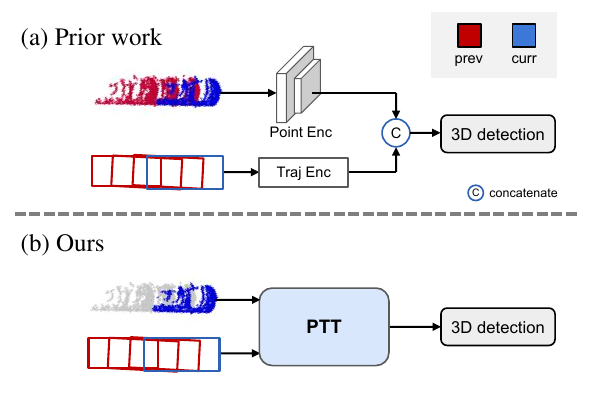}
    \caption{
    \textbf{Different approaches for temporal 3D object detection.
    } 
    (a) Existing methods~\cite{mppnet, msf} require per-frame point clouds as input, resulting in more memory overhead. In addition, the straightforward concatenation of point and trajectory features overlooks their interactions across features.
    (b) Our approach minimizes the space requirement for the memory bank by utilizing only the current frame's point cloud as input. Moreover, we introduce a point-trajectory transformer (PTT) to effectively integrate point and trajectory features. Note that the {\color{gray}gray} point clouds indicate that they are not utilized).
    }
    \label{fig:example}
    \vspace{-11pt}
\end{figure}

One straightforward method to tackle sequential data is concatenating all multi-frame LiDAR as input~\cite{Zhou_centerformer, 3d-man}, followed by the original 3D detection pipeline, which provides performance benefit via the denser information. To better model the information of objects instead of the whole point cloud, recent methods~\cite{mppnet, msf} typically adopt a two-stage approach. In the first stage, they generate bounding box candidates and select objects' point clouds using region proposal networks. 
Subsequently, these approaches concatenate point features with proposal features for 3D object detection, as illustrated in Figure \ref{fig:example}(a).

Nevertheless, these proposal-based techniques suffer from two main limitations. 
First, sampling object point clouds for each frame in sequential modeling causes a memory burden for a long sequence during the training and testing phases. 
Second, these methods focus on the design of networks for point cloud sequences, which often neglects the significance of the interaction between point and proposal features.
In this paper, we propose PTT, a Point-Trajectory Transformer for efficient temporal 3D object detection to handle these two challenges (see Figure \ref{fig:example}(b)). 

We emphasize the significance of the multi-frame proposal trajectory. Instead of storing sampled point clouds from all previous LiDARs, we only input the current-frame object point cloud. We employ PTT to model the relationship between single-frame point clouds and multi-frame proposals, facilitating efficient utilization of rich multi-frame LiDAR data with reduced memory overhead.
Our motivation is recognizing that previous proposals can inherently approximate the object's information without using the entire object point cloud. Consequently, we can effectively leverage historical information without requiring more memories.
While previous methods necessitate a large memory bank to store point cloud sequences, our pipeline processes concise information from more frames with limited memory usage. This enables PTT to capture object trajectories over a longer time duration (\eg, 64 frames for ours \vs 8 frames for MSF~\cite{msf}).

Furthermore, unlike previous methods that concatenate proposal trajectories with point cloud features, we introduce the point-trajectory transformer with long short-term memory to model the relationship between single-frame point clouds and multi-frame proposals. In this approach, we partition information obtained from multiple frames into long-term and short-term encoding to enhance representations over temporal information. In addition, we infer the future possible trajectory of the proposals to generate future-aware point features. Finally, we design an aggregator to facilitate interactions between point and trajectory features. 

We conduct extensive experiments on the Waymo Open Dataset~\cite{Sun_2020_CVPR} to show the efficiency and effectiveness of the proposed PTT. For example, our method can achieve favorable performance against state-of-the-art schemes, using longer frames with lower memory requirements. Moreover, we present detailed ablation studies and analyses to demonstrate the usefulness of proposed components in PTT.
The main contributions of this work are: 
\begin{compactitem}
    \item We propose a temporal 3D object detection framework by efficiently considering only the point cloud in the current frame, along with longer historical proposal trajectories.
    \item We introduce a point-trajectory transformer with long short-term memory and future encoding that efficiently extracts features from single-frame point clouds and multi-frame trajectories, eliminating the need for storing multi-frame point clouds. 
    \item We present comprehensive results with analysis to demonstrate the effectiveness of our method, allowing longer temporal information with lower cost while still achieving favorable results against other approaches. 
\end{compactitem}

%% file: sec/1_RelatedWork.tex
\section{Related Work}

\noindent {\bf Single-Frame 3D Object Detection.}
Recent 3D object detection methods mainly focus on learning effective feature representations of point clouds. They can be categorized into two categories: point-based and voxel-based approaches.
Point-based techniques \cite{shi2019pointrcnn,QiLHG19,YangS0SJ19,ZhangHXMWG22} directly operate on point clouds in continuous space, extracting semantic and geometric features. 
In~\cite{shi2019pointrcnn}, PointRCNN presents a two-stage approach, initially segmenting foreground points and generating object-wise proposals, followed by refinement in canonical coordinates. 3DSSD~\cite{luo2021m3dssd} introduces a fusion sampling technique utilizing feature distance to ensure comprehensive information preservation. Point-GNN~\cite{Point-GNN} uses a graph neural network to create a more condensed representation of point cloud data. Furthermore, Pointformer~\cite{Pan_2021_CVPR} introduces a transformer module that incorporates local and global attention, operating directly on point clouds.

Voxel-based methods \cite{yan2018second,VoxelNet,yin2021center} rasterize point clouds into fixed-size spatial voxels and use 2D or 3D CNNs to extract dense features. 
Second~\cite{yan2018second} introduces a sparse convolution method to enhance the efficiency of point cloud processing.
VoxelNet~\cite{VoxelNet} encodes voxel features from the raw point cloud using a dense region proposal network for 3D object detection. In~\cite{mao2021voxel}, Voxel-RCNN employs voxel Region-of-Interest (RoI) pooling to extract voxel features within proposals, facilitating subsequent refinement.
Moreover, CenterPoint~\cite{yin2021center} transforms the sparse output from a backbone network into a dense Bird's Eye View (BEV) feature map. The model predicts a dense heatmap representing the center locations of objects using the information from this dense feature map.
Our approach uses CenterPoint~\cite{yin2021center} as the proposal generation network to predict 3D proposals for each frame. Subsequently, we introduce a temporal transformer to aggregate points and trajectory features for refining the proposals effectively.

\begin{figure*}[t]
\centering

\includegraphics[width=0.98\textwidth]{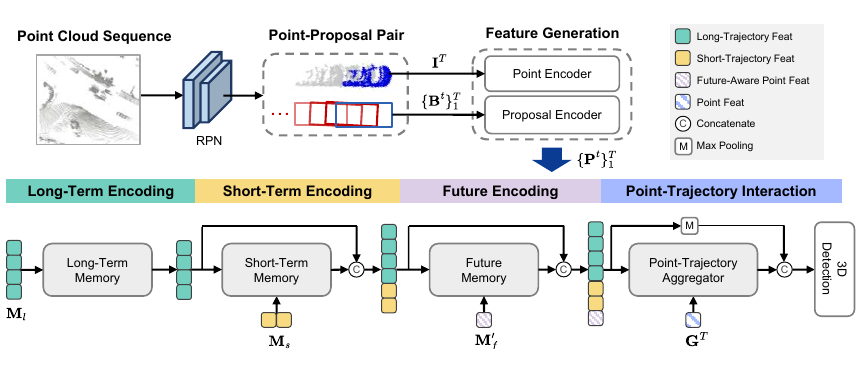} 
\caption{\textbf{Overall framework of the proposed Point-Trajectory Transformer (PTT).} 
First, we utilize a region proposal network (RPN) at timestamp $T$ to generate proposals $\mathbf{B}^T$ for each frame, sample the corresponding point-of-interest $\mathbf{I}^T$, and connect past $T$-frame 3D proposals to form proposal trajectories $\{\mathbf{B}^1,...,\mathbf{B}^T\}$.
Then, we take the single-frame point cloud for each object and its previous multi-frame trajectory as input to generate point-trajectory features ${\mathbf P}^t$, which avoid storing per-frame points to mitigate memory bank's overhead (Section \ref{sec:feat}).
Finally, we present a point-trajectory transformer (PTT) to fuse features, which consists of four components: Long-term and Short-term encoders for extracting two types of features, a future-aware module for extracting future-aware point features, and a point-trajectory aggregator for adaptive interaction between trajectory and current frame's point features ${\mathbf G}^T$ (Section \ref{sec:ptt}).
}
\label{fig:arch}
\vspace{-20pt}
\end{figure*}

\smallskip \noindent {\bf Temporal 3D Object Detection.}
Compared with 3D object detectors solely from a single frame, including temporal information in multi-frame scenarios~\cite{mppnet, msf, 3d-man, fandf} yields a more dense and comprehensive understanding of the surrounding environment. 
Fast-and-furious~\cite{fandf} concatenates hidden feature maps of multi-frame point clouds to aggregate temporal information. 
3D-MAN~\cite{3d-man} utilizes an attention mechanism to align 3D objects from various views and employs a memory bank to store and aggregate temporal information for processing lengthy point cloud sequences.
Furthermore, Offboard3D~\cite{Offboard3d} and MPPNet~\cite{mppnet} associate the detected boxes from individual frames of the sequence as proposal trajectories and extract high-quality proposal features by sampling sequential point clouds along these trajectories to enhance 3D detection performance. 
In CenterFormer~\cite{Zhou_centerformer}, features of multi-frame objects are fused with object queries through cross-attention operations. On the other hand, MSF~\cite{msf} generates proposals in the current frame and propagates them to explore features in earlier frames, offering efficiency and suitability for online detection systems. Nevertheless, the backward propagation process still necessitates a substantial memory bank to store the point cloud sequence, which may be less practical for on-device deployment.

Despite promising advances, most methods mentioned above sample objects' point clouds or the entire scene's point clouds in each frame, which may demand a large memory bank to store, especially when dealing with long sequences. In contrast, our proposed PTT achieves memory-efficient online detection by employing a point-trajectory transformer to fuse information from the current frame's point cloud and multi-frame proposal trajectory.

%% file: sec/2_Approach.tex
\section{Proposed Approach}
\subsection{Framework Overview}
\label{sec:approach}

In this paper, we propose a Point-Trajectory Transformer (PTT) for efficient temporal 3D object detection.
The overall framework is shown in Figure~\ref{fig:arch}. 
We employ a region proposal network (RPN) to produce proposals for each frame, sample the corresponding points of interest based on the proposals, and establish connections among past $T$-frames' 3D proposals to construct box trajectories.
%
To mitigate the memory bank's overhead, our approach stores solely the single-frame point cloud and multi-frame proposals (see further analysis in Section~\ref{sec:memory}).

We first encode two types of features, point-to-proposal and proposal-to-proposal features, by sampling the current frame's point cloud and multi-frame trajectories (Section~\ref{sec:feat}).
Then, we present our proposed point-trajectory transformer to fuse sequential features, which mainly consists of the following modules: 
1) Long-term and short-term memory modules are designed to encode two distinct types of sequential features. 2) A future encoder is employed to predict the object's future trajectory, facilitating the encoding of future-aware features. 3) We introduce a point-trajectory aggregator to integrate an object's current-frame point features with its trajectory features (Section~\ref{sec:ptt}). 
 
\subsection{Analysis of Memory Bank's Size}
\label{sec:memory}
Recent state-of-the-art temporal 3D object detectors often adopt a two-stage approach \cite{mppnet, msf}. In the initial stage, they generate 3D box candidates using the region proposal network and subsequently sample points based on these boxes for integrating information across temporal frames.
However, these approaches need to store either extensive raw point cloud data~\cite{msf} or point clouds of identified objects~\cite{mppnet} based on the length of frames. This leads to potential memory issues, especially with a large number of frames.

We compare our approach with MPPNet~\cite{mppnet} and MSF~\cite{msf} regarding the space complexity of the memory bank in Table~\ref{tab:memory}. Assuming that each object requires $\mathcal{O}(N)$ for storing its point clouds and $\mathcal{O}(O)$ for the proposal in each frame, where the average number of objects in the scene is $k$. Here, $\mathcal{O}(F)$ denotes the average space complexity for the per-frame point cloud input.
For the region proposal network, we employ CenterPoint~\cite{yin2021center} with an input of 4 frames of accumulated point clouds. This implies that all the approaches require $\mathcal{O}(4F)$ space complexity to deal with the input. 

Regarding the point cloud of objects, MPPNet~\cite{mppnet} with 16 frames (default setting) needs to store each object's per-frame proxy points $N'=N/2$, so the memory complexity is $\mathcal{O}(16kN')={O}(8kN)$, which will increase the complexity based on the frame number.
For MSF~\cite{msf} with 8-frames input, the memory complexity for dealing with the point clouds is $\mathcal{O}(8F)$ since they need to keep the entire sequential point cloud for propagating the proposals into all previous frames to find previous bounding box candidates.
Instead, our approach only samples the current-frame object point cloud, which only requires $\mathcal{O}(kN)$ space complexity without being proportional to the frame number, allowing us to leverage longer frames (e.g., 64).

For the complexity of proposals, MPPNet, MSF, and our PTT are $\mathcal{O}(16kO)$, $\mathcal{O}(kO)$, and $\mathcal{O}(64kO)$, respectively, where MSF's complexity is not proportional to the frame number.
Although our space complexity for the proposal part is more significant due to the use of 64 frames, compared with the point cloud, the memory bank size for the proposal is much smaller.
For instance, each object stores a point cloud with 128 points $\times$ 4 dimensions, including 3D location and intensity. In contrast, a proposal for each object only requires 9 dimensions, encompassing the 3D center point, 3D size, heading, and 2D BEV velocity.

Assuming there are $k=200$ objects and the average point number in the point clouds is $160,000$ for each frame, we can calculate that $\mathcal{O}(F)$ and $\mathcal{O}(kN)$ are approximately 355$\times$ and 55$\times$ larger than $\mathcal{O}(kO)$, respectively.
We can then obtain that the memory bank size of our approach is the lowest,
which indicates that our approach efficiently minimizes the memory bank's storage requirement while maximizing the leveraged frame number.

\subsection{Point-Trajectory Feature Generation}
\label{sec:feat}
Unlike the prior work~\cite{mppnet, msf} keeping per-frame objects or entire point clouds, we advocate using single-frame points and multi-frame proposals, a more storage-efficient solution as discussed in Section~\ref{sec:memory}. 
We follow MPPNet~\cite{mppnet} to infer the previous location of the bounding box based on the estimated velocity for all proposals to link cross-frame boxes.
%
After that, we propose to encode two kinds of features: Point-to-Proposal and Proposal-to-Proposal features.

\smallskip \noindent {\bf Point-to-Proposal Features.}
Consider the current frame's points-of-interest ${\mathbf I}^T$ and the latest $T$-frame box trajectories $\{{\mathbf B^{1}},...,{\mathbf B^T}\}$ for a specific object, our objective is to encode the point features based on historical proposals.
We first generate point-to-proposal features based on the offset between points and points of the bounding box (i.e., center and eight corners) $\{{{\mathbf b}^{t}_{i} \in {\mathcal C}_3({\mathbf B^t})}: i=0,...,8\}$~\cite{ct3d}: 
\begin{equation}
\begin{aligned}
    {\mathbf G}^t =\text{MLP}(\text{Concat}(\{{\mathbf I}-{\mathbf b}^{t}_{i}\}_{i=0}^8, \Delta t)) \in \mathbb{R}^{N \times C},
\label{eq:point}
\end{aligned}
\end{equation}
where $N$ is number of the sampled points, $C$ is the feature dimension, and $\Delta t$ is the time difference between ${\mathbf b}^t$ to the current frame. Then $\{{\mathbf G}^t\}^T_1$ denotes the sequential Point-to-Proposal features that encode the point features based on its proposal at different frame $t$. To this end, we are not required to store each frame's point cloud for the object but can still preserve its geometric and motion features related to previous frames.

\smallskip \noindent {\bf Proposal-to-Proposal Features.}
By considering proposals from the previous $T$ frames, we can encode the relative location of the proposals compared to the current frame, enhancing the model's ability to capture the motion pattern more effectively. The Proposal-to-Proposal features can be defined as:
\begin{equation}
    {\mathbf F}^t =\text{MLP}(\text{Concat}({\mathbf B}^t_{xyz}-{\mathbf B}^T_{xyz}, {\mathbf B}^t_{whl\theta},\Delta t)) \in \mathbb{R}^{C},
\label{eq:proposal}
\end{equation}
where ${\mathbf B}_{xyz}$ and ${\mathbf B}_{whl\theta}$ represent the attributes of the proposal, including the center location $(x, y, z)$ and the size of the bounding box $(w, l, h)$ respectively, along with the heading $\theta$.
To this end, $\{{\mathbf F}^t\}^T_1$ is the sequential Proposal-to-Proposal features.

\input{tab/memory}

\smallskip \noindent {\bf Sequential Point-Trajectory Features.}
After deriving the Point-to-Proposal and Proposal-to-Proposal features from \eqref{eq:point} and \eqref{eq:proposal}, we can combine them into integrated features.
Specifically, we compress the per-frame Point-to-Proposal features using a MaxPool operation and concatenate them with the corresponding Proposal-to-Proposal features to acquire per-frame point-trajectory features:
\begin{equation}
    {\mathbf P}^t = \text{MLP}(\text{Concat}({\mathbf F}^t, \text{MaxPool}({\mathbf G}^t))).
\label{eq:total}
\end{equation}
In this manner, we can use ${\{\mathbf P}^t\}^T_1$ to represent the sequential point-trajectory information, which is utilized for further feature learning.

\subsection{Point-Trajectory Transformer}
\label{sec:ptt}
After acquiring the sequential point-trajectory features, our objective is to devise a point-trajectory transformer with long short-term memory to encode these features. Our transformer comprises four modules: long-term memory, short-term memory, future-aware encoder, and point-trajectory aggregator module.

\smallskip \noindent {\bf Long-Term Memory.}
Contrary to prior approaches that may be limited to handling short sequences, our memory-efficient solution can manage significantly longer sequences. We partition the feature sequence into long-term and short-term memories to capture sequential information effectively across various periods.
Considering the total frame length $T$, we separate the sequential point-trajectory features ${\{\mathbf{P}^t}\}$ into the short memory as $\mathbf{M}_s = {\{\mathbf{P}^T,...,\mathbf{P}^{T-{\mathbf m}_s+1}}\}$ with a short frame number ${\mathbf m}_s$, and $\mathbf{M}_l = {\{\mathbf{P}^{{\mathbf m}_l},...,\mathbf{P}^{1}}\}$ with a long frame number ${\mathbf m}_l$, where ${\mathbf m}_l + {\mathbf m}_s = T$.

For the long-term memory ${\mathbf M}_l \in \mathbb{R}^{{\mathbf m}_l\times C}$, we utilize the multi-head self-attention encoder layers to encode the long-term features:
\begin{align}
  &  \mathbf{Q} = {\mathbf M}_l \mathbf{W}_q, \mathbf{K} = {\mathbf M}_l \mathbf{W}_k, \mathbf{V} = {\mathbf M}_l \mathbf{W}_v, \nonumber \\
  &\hat{{\mathbf M}}_l ={\rm FFN}({\rm{MultiHead}}(\mathbf{Q},\mathbf{K},\mathbf{V})),
\label{eq-att-4}
\end{align}
where $\mathbf{W}_*$ denotes the learnable parameters for the long-term memory.
We utilize one linear layer followed by the ReLU activation to construct our feed-forward network $\rm{FFN}$ (see  \cite{vaswani2017SA} for details about the self-attention layer $\rm{MultiHead}$). To this end, we can obtain the aggregated long-term trajectory features $\hat{{\mathbf M}}_l$.

\smallskip \noindent {\bf Short-Term Memory.} 
Short-term features are crucial for learning 3D bounding boxes in the current frame, as they capture the dynamic status close to the current frame.
We use the short-term memory ${\mathbf M}_s$ as a query to extract valuable trends in historical sequences from the aggregated long-term memory $\hat{{\mathbf M}}_l$.
We exploit the transformer module to model the relationship between short- and long-term memories:
\begin{align}
  &  \mathbf{Q}^s = {\mathbf M}_s \mathbf{W}_q^s, \mathbf{K}^l = \hat{{\mathbf M}}_l \mathbf{W}_k^l, \mathbf{V}^l = \hat{{\mathbf M}}_l \mathbf{W}_v^l, \nonumber \\
  &\hat{{\mathbf M}}_s ={\rm FFN}({\rm{MultiHead}}(\mathbf{Q}^s,\mathbf{K}^l,\mathbf{V}^l)),
\label{eq-att-5}
\end{align}
where $\hat{{\mathbf M}}_s$ denotes the interactive short-term memory features. 
Then, we connect one FC layer to learn the residual offset between the current frame's proposal and the corresponding ground truth bounding box. This ensures that the learned short-term features can represent short historical motion cues.

\begin{figure}[!t]
    \centering
    \includegraphics[width=1\linewidth]{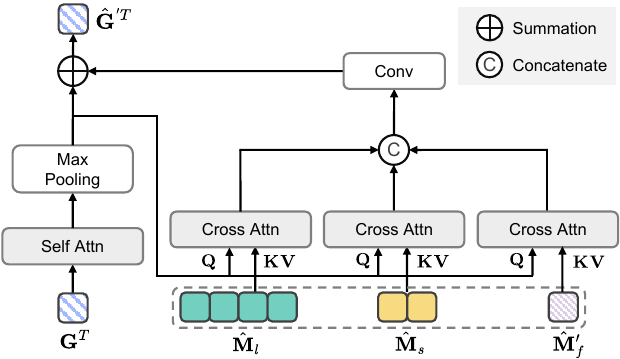}
    \caption{
    \textbf{
    Point-Trajectory Aggregator.
    } 
    Current-frame point cloud features $\mathbf{G}^T$ are squeezed and interact with long-term memory $\hat{{\mathbf M}}_l$, short-term memory $\hat{{\mathbf M}}_s$, and future memory $\hat{{\mathbf M}}'_f$.
    See Section~\ref{sec:ptt} for more details.
    }
    \label{fig:pt_agg}
    \vspace{-3mm}
\end{figure}

\smallskip \noindent {\bf Future-Aware Activation.}
Inspired by the anticipation of future action in the action recognition field ~\cite{onlineaction2019iccv}, we observe that future information is also important for 3D estimation in the current frame by improving feature learning.
Considering a 3D proposal box with $(x,y,z,w,l,h,\theta)$ with the estimated velocity $(v_x, v_y)$ at frame $T$, we can infer the box's state at future time difference $\Delta t$ as $\mathbf{B}^{T+\Delta t} = (x+\Delta t \cdot v_x, y+\Delta t \cdot v_y, z, w, l, h, \theta + \Delta t \cdot v_{\theta})$, where $v_{\theta}$ can be calculated by the difference within previous frames: $v_{\theta} = (\theta_{T} - \theta_1)/T$. 

Next, we can utilize \eqref{eq:total} to derive the future-aware features $\mathbf{M}_f = {\{\mathbf{P}^{T+\mathbf{m}_f},...,\mathbf{P}^{T+1}}\}$ for $\mathbf{m}_f$ future frames based on the inferred future box ${\{\mathbf{B}^{T+t}}\}^{t=\mathbf{m}_f}_{t=1}$.
We then concatenate the long-term $\hat{{\mathbf M}}_l$ and the short-term memory $\hat{{\mathbf M}}_s$ to generate the whole memory features ${\mathbf M} = [\hat{{\mathbf M}}_l,\hat{{\mathbf M}}_s]$, where $[\cdot,\cdot]$ means the feature concatenation operation along the temporal dimension. 
Afterward, since the inferred future boxes may be inaccurate, we concatenate the learnable queries $\mathbf{q}_{f} \in \mathbb{R}^{\mathbf{m}_f \times C}$ to the future features along the channel dimension, followed by the convolution operation: $\mathbf{M}_f'=\text{Conv}(\text{Concat}(\mathbf{q}_{f}, \mathbf{M}_f))$, facilitating additional learning for potential correction.
Then the concatenated features $\mathbf{M}_f'$ are fed to the multi-head self-attention encoder layers:
\begin{align}
  &  \mathbf{Q}^f = \mathbf{M}_f'\mathbf{W}_q^f, \mathbf{K}^f = \mathbf{M}\mathbf{W}_k^f, \mathbf{V}^f = \mathbf{M}\mathbf{W}_v^f, \nonumber \\
  &  \hat{\mathbf{M}}_f'={\rm FFN}({\rm{MultiHead}}(\mathbf{Q}^f,\mathbf{K}^f,\mathbf{V}^f)),
\label{eq-att-6}
\end{align}
where $\hat{\mathbf{M}}_f'$ is the future-aware features. Similar to the short-term memory encoder, we connect the feature to one FC layer that learns the residual offset,
aiding the model in capturing meaningful representations.

\input{tab/waymo_val}

\smallskip \noindent {\bf Point-Trajectory Aggregator.}
After obtaining the enhanced sequential point-trajectory features based on long-term, short-term, and future encoding, we integrate them with the current frame's point features into the proposed point-trajectory aggregator as shown in Figure~\ref{fig:pt_agg}.  

For the current frame's point cloud features $\mathbf{G}^T \in \mathbb{R}^{N \times C}$, we first employ a self-attention encoding block, followed by the max-pooling operation to obtain the compressed point features $\hat{\mathbf{G}}^T \in \mathbb{R}^{C}$.
Subsequently, three distinct cross-attention modules are used to interact with the trajectory memories: long-term, short-term, and future-aware memories, with point features serving as queries to generate three different queried memories:
$\hat{\mathbf{M}}_{lp}$,  $\hat{\mathbf{M}}_{sp}$, and $\hat{\mathbf{M}}_{fp}$, where ${\mathbf{M}}_{*p}$ indicates the point-interacted features for long-term, short-term, and future encoding, respectively.

After that, we concatenate these three features and apply a convolution layer, followed by the summation of the point features to obtain the final integrated point features:
\begin{equation}\hat{\mathbf{G}}^{'T}=\hat{\mathbf{G}}^T+\text{Conv}([\hat{\mathbf{M}}_{lp},\hat{\mathbf{M}}_{sp},\hat{\mathbf{M}}_{fp}])),
\end{equation}
where $\hat{\mathbf{G}}^{'T}$ is the enhanced point features for the current frame.
Furthermore, we apply the Max Pooling operation to obtain final trajectory features ${\mathbf M}_{\text{Traj}}=\text{MaxPool}([\hat{{\mathbf M}}_l,\hat{{\mathbf M}}_s, \hat{{\mathbf M}}'_f])$ and concatenate it with the enhanced point features $\hat{\mathbf{G}}^{'T}$ for the following 3D box learning.

\smallskip \noindent {\bf Loss Functions.} 
Finally, the learned features are sent to the detection head for 3D box estimation.
The overall detection loss $\mathcal{L}_{\mathrm{det}}$ comprises the confidence prediction loss $\mathcal{L}_{\mathrm{conf}}$ and the box regression loss  $\mathcal{L}_{\mathrm{reg}}$, formulated as: 
\begin{equation}
\mathcal{L}_{\mathrm{det}}=\mathcal{L}_{\mathrm{conf}}+\alpha\mathcal{L}_{\mathrm{reg}},
\end{equation}
where $\alpha$ represents the weight for balancing two losses. Our confidence prediction loss $\mathcal{L}_{\mathrm{conf}}$ and box regression loss $\mathcal{L}_{\mathrm{reg}}$ are implemented using the binary cross-entropy loss and box regression loss respectively, similar to those employed in CT3D~\cite{ct3d}.

%% file: tab/memory.tex
\setlength{\tabcolsep}{0.029\linewidth}{
\begin{table}[t]
    \footnotesize
    \centering
    \begin{tabular}{lc|ccc}  
        \hline
        \multirow{2}{*}{Method}& \multirow{2}{*}{Frame Num} & \multicolumn{3}{c}{Space Complexity} \\
         & & RPN & Point & Proposal \\
        \hline\hline
        MPPNet~\cite{mppnet} & 16 & $\mathcal{O}(4F)$ & $\mathcal{O}(8kN)$ & $\mathcal{O}(16kO)$ \\ 
        MSF~\cite{msf}    & 8  & $\mathcal{O}(4F)$ & $\mathcal{O}(8F)$ & $\mathcal{O}(kO)$ \\
        PTT (Ours)  & 64 & $\mathcal{O}(4F)$ & $\mathcal{O}(kN)$ & $\mathcal{O}(64kO)$\\
        \hline
    \end{tabular}
    \vspace{-1mm}
    \caption{\textbf{Comparisons with various methods regarding space complexity.} We utilize CenterPoint~\cite{yin2021center} with 4-frame input as the region proposal network. $F$ and $N$ denote the numbers representing the whole point cloud and individual objects, respectively. $k$ is the average object number in a scene. $O$ represents the dimension of the proposal. See Section~\ref{sec:memory} for more details.
    }
    
    \label{tab:memory}
\end{table}
}

%% file: tab/waymo_val.tex
\begin{table*}[!t]
\centering
\resizebox{1\linewidth}{!}{
\begin{tabular}{l|c|cc|cc|cc|cc}
\hline
& & \multicolumn{2}{c|}{ALL (3D mAPH)} & \multicolumn{2}{c|}{VEH (3D AP/APH)} & \multicolumn{2}{c|}{PED(3D AP/APH)} & \multicolumn{2}{c}{CYC(3D AP/APH)} \\ \cline{3-10} 
\multirow{-2}{*}{Method}   &    \multirow{-2}{*}{Frames} & L1  & L2 & L1 & L2  & L1 & L2 & L1  & L2  \\ \hline\hline
SECOND~\cite{yan2018second} & 1  & 63.05   & 57.23   & 72.27/71.69   & 63.85/63.33       &  68.70/58.18 & 60.72/51.31         & 60.62/59.28     &58.34/57.05 \\
IA-SSD \cite{zhang2022not} & 1& 64.48&58.08&70.53/69.67& 61.55/60.80 &69.38/58.47& 60.30/50.73& 67.67/65.30 &64.98/62.71\\
PointPillar~\cite{Lang2019pointpillars}           & 1             & 63.33      & 57.53       & 71.60/71.00  & 63.10/62.50        &  70.60/56.70   & 62.90/50.20     & 64.40/62.30    & 61.90/59.90                 \\
LiDAR R-CNN~\cite{li2021lidar}           & 1                       & 66.20       & 60.10      & 73.50/73.00       & 64.70/64.20      & 71.20/58.70       & 63.10/51.70      & 68.60/66.90      & 66.10/64.40     \\
RSN~\cite{rsn}                     & 1 & -  & -     & 75.10/74.60         & 66.00/65.50        & 77.80/72.70                 & 68.30/63.70                 & -                 &  -               \\
PV-RCNN~\cite{shi2020pv}             & 1  & 69.63   & 63.33   &77.51/76.89 & 68.98/68.41  & 75.01/65.65  & 66.04/57.61                  & 67.81/66.35   & 65.39/63.98                 \\
Part-A2~\cite{shi2020parta2}            & 1                      & 70.25  & 63.84   & 77.05/76.51   & 68.47/67.97  & 75.24/66.87     & 66.18/58.62     & 68.60/67.36  & 66.13/64.93         \\
VoTR~\cite{mao2021voxel} & 1 & - & - & 74.95/74.25 &65.91/65.29 & - & - & - & - \\
SWFormer-1f\cite{swformer} &1& -& -&77.8/77.3& 69.2/68.8&80.9/72.7 &72.5/64.9 &-&- \\
PillarNet \cite{shi2022pillarnet}& 1& 74.60&68.43&79.09/78.59& 70.92/70.46& 80.59/74.01& 72.28/66.17& 72.29/71.21& 69.72/68.67\\
PV-RCNN++~\cite{pvrcnnplus}    & 1     & 75.21  & 68.61   & 79.10/78.63       & 70.34/69.91 & 80.62/74.62  & 71.86/66.30	 & 73.49/72.38  & 70.70/69.62                 \\ 
FSD~\cite{fan2022fully}        & 1     & 77.4 & 70.8 & 79.2/78.8 & 70.5/70.1 & 82.6/77.3 & 73.9/69.1 & 77.1/76.0 & 74.4/73.3\\

\hline
3D-MAN~\cite{3d-man} & 16                      &-                   &-     & 74.53/74.03  & 67.61/67.14 & -  & -                & -                &  -               \\
SST-3f~\cite{sst} & 3 & - & - & 78.66/78.21 & 69.98/69.57 & 83.81/80.14 & 75.94/72.37 & - & - \\
SWFormer-3f~\cite{swformer} & 3 & - & - & 79.4/78.9 & 71.1/70.6 & 82.9/79.0 & 74.8/71.1 & - & - \\
CenterFormer~\cite{Zhou_centerformer} & 4 & 77.0 & 73.2 & 78.1/77.6 & 73.4/72.9 & 81.7/78.6 & 77.2/74.2 & 75.6/74.8 & 73.4/72.6 \\
CenterFormer~\cite{Zhou_centerformer} & 8 & 77.3 & 73.7 & 78.8/78.3 & 74.3/73.8 & 82.1/79.3 & 77.8/75.0 & 75.2/74.4 & 73.2/72.3\\
MoDAR~\cite{modar} & 92 & - & - & 81.0/80.5 & 73.4/72.9 & 83.5/79.4 & 76.1/72.1 & - & - \\
MPPNet~\cite{mppnet}   & 4                       & 79.83   & 74.22  & 81.54/81.06 & 74.07/73.61 &84.56/81.94    & 77.20/74.67      & 77.15/76.50     & 75.01/74.38                 \\
MPPNet~\cite{mppnet} & 16                      & 80.40  & 74.85   & 82.74/82.28 & 75.41/74.96  & 84.69/82.25  & 77.43/75.06      &77.28/76.66    & 75.13/74.52 \\  
MSF~\cite{msf} & 4 & 80.20 & 74.62 & 81.36/80.87 & 73.81/73.35 & 85.05/82.10 & 77.92/75.11 & 78.40/77.61 & 76.17/75.40 \\ 
MSF~\cite{msf} & 8 & 80.65 & 75.46 & 82.83/82.01 & 75.76/75.31 & 85.24/82.21 & 78.32/75.61 & \textbf{78.52}/77.74 & \textbf{76.32}/\textbf{75.47}\\ \hline 
PTT (Ours) & 32 & 81.11 & 75.48 & 83.31/82.82 & 75.83/75.35 & 85.91/82.94 & 78.89/76.00 & 78.29/77.57 & 75.81/75.11   \\
PTT (Ours) & 64 & \textbf{81.32} & \textbf{75.71} & \textbf{83.71/83.21} & \textbf{76.26/75.78} & \textbf{85.93/82.98} & \textbf{78.90}/\textbf{76.02} & 78.51/\textbf{77.79} & 76.01/75.32   \\
\end{tabular}}
\vspace{-1mm}
\caption{\textbf{Performance comparisons on the validation set of the Waymo Open Dataset.} The metrics are 3D AP and APH for both L1 and L2 difficulties. We use \textbf{bold} numbers to highlight the best results.}
\label{table:waymo_val}
\end{table*}

%% file: sec/3_Experiments.tex
\section{Experiments}
\subsection{Experimental Setup}
\label{sec:setup}
\noindent{\bf {Dataset and Evaluation Metrics.}}
We evaluate the proposed PTT method on the Waymo Open Dataset (WOD)~\cite{Sun_2020_CVPR}. 
WOD comprises 1,150 sequences, partitioned into 798 training, 202 validation, and 150 testing sequences. Each sequence spans 20 seconds and is captured by a 64-line LiDAR sensor at 10Hz. Evaluation of the WOD dataset employs mean average precision (mAP)and mAP weighted by heading accuracy (mAPH) as metrics. The dataset includes three object categories: ``Vehicle,'' ``Pedestrian'', and ``Cyclist''. These classes are further divided into two difficulty levels, LEVEL1 and LEVEL2. LEVEL1 pertains to objects with more than 5 points, while LEVEL2 includes objects with fewer than 5 points but at least 1 point.

\smallskip\noindent{\bf Implementation Details.}
We utilize the CenterPoint~\cite{yin2021center} as our region proposal network (RPN) to generate first-stage high-quality 3D proposals, which take four adjacent frames at input and add one additional head for object velocity prediction similar to \cite{mppnet, msf}.
Building upon these proposals, we train our proposed PTT for 6 epochs, utilizing the ADAM optimizer with an initial learning rate of 0.003 and a batch size of 4.
We set an IoU matching threshold of 0.5 for generating the proposed trajectories in the first stage. Also, we apply proposal-centric box jitter augmentation to the per-frame 3D proposal box, following the approach outlined in PointRCNN~\cite{shi2019pointrcnn}.
Each proposal is randomly sampled with 128 raw LiDAR points.
The feature dimension for point clouds and trajectories is configured as 128. 
We set $\alpha$ as 2 following CT3D~\cite{ct3d} to balance the overall loss terms.
During training, intermediate supervision is implemented by adding a loss to the output of each learning block, and all the intermediate losses are summed to train the model. At the testing stage, only bounding boxes and confidence scores predicted from the last block are utilized.
All the experiments are implemented in Pytorch on an NVIDIA 3090 GPU.
Following MPPNet~\cite{mppnet} to reduce computational cost, we conduct ablation studies by using a light-cost setting of training our PTT with 32 frames for 3 epochs.
%
More details and results are presented in the supplementary materials.

\subsection{Main Results}

\smallskip\noindent{\bf Comparisons with single-frame methods.}
In Table~\ref{table:waymo_val}, we compare our proposed PTT with different state-of-the-art methods on the Waymo validation set. 
First, in comparison to state-of-the-art single-frame methods, e.g., FSD~\cite{fan2022fully}, our PTT with 64-frame sequences notably enhances the overall 3D mAPH (LEVEL 2), achieving improvements of 5.6\%, 6.9\%, and 2.0\% on vehicles, pedestrians, and cyclists categories, respectively.
%

\smallskip\noindent{\bf Comparisons with multi-frame methods.}
Second, our approach performs favorably against several multi-frame methods in most evaluation metrics.
Compared with two top-performing multi-frame methods, MPPNet~\cite{mppnet} and MSF~\cite{msf}, our approach consistently performs better in 11 out of 14 evaluation metrics with the same region proposal network~\cite{yin2021center}.

\input{tab/abl_longshort}

\smallskip\noindent{\bf Runtime Performance.}
Specifically, our PTT approach, employing 64 frames, performs better than MPPNet~\cite{mppnet} using 16 frames, while consuming less memory overhead (see Table~\ref{tab:memory}) since MPPNet needs to store per-frame sampled object point clouds. 
In addition, MPPNet introduces an MLP-Mixer module to interact points among sequences, incurring more computational overhead. In contrast, our point-trajectory transformer efficiently handles long sequential data. For instance, while MPPNet takes approximately 1100ms to process 16-frame sequential data, our PTT only requires about 150ms to run 64-frame information on the same devices. This efficiency demonstrates the effectiveness of our approach.

Furthermore, when comparing our approach using 64 frames with MSF~\cite{msf} using 8 frames, our method also achieves better performance across most metrics. It is worth mentioning that, although our PTT exhibits a comparable runtime to MSF (approximately 160ms), their approach necessitates retaining the entire sequential point cloud to propagate the current frame's proposals into previous frames. This requirement results in a larger memory bank, potentially causing memory overhead and limiting its applicability to longer sequences.

\subsection{Ablation Study}
\label{sec:ablation}
\noindent{\bf Importance of long-term and short-term encoding.}
We show the ablation study of our long-term and short-term encoding in Table~\ref{tab:abl_longshort}.
Overall, using both encodings yields the most favorable results, which shows the complementary properties of these two memory encoding ways.
The reason is that the long-term memory allows the model to capture the long-range dependencies, while the short-term memory focuses on the recent state, capturing immediate context. Adding both provides a more comprehensive view of temporal patterns.

\smallskip{\noindent{\bf Effectiveness of the proposed future encoding.}}
As demonstrated in Table~\ref{abl_future}, the inclusion of additional future encoding enhances the performance of our proposed PTT, leading to an increase of 0.32\%, 0.49\%, and 0.42\% in APH across vehicle, pedestrian, and cyclist categories within the Level 2 difficulty.
This observation validates the value of latent information from future timestamps, aiding the model in contributing to the learning process via the availability of longer trajectory information spanning from the past to the near future.

\input{tab/abl_future}

\smallskip{\noindent{\bf Effectiveness of the proposed Point-to-Proposal and Proposal-to-Proposal features.}}
We investigate the effectiveness of the proposed point-to-proposal and proposal-to-proposal features in Table~\ref{tab:abl_ppfeat}. We show that jointly applying both features achieves the optimal L1 (mAP/mAPH) and L2 (mAP/mAPH) scores. Utilizing proposal-to-proposal features alone is better than using point-to-proposal features, as the information of the proposals provides more comprehensive information for representing an object.

\smallskip{\noindent{\bf Effectiveness of the proposed point-trajectory transformer.}}
We demonstrate the effectiveness of our proposed point-trajectory transformer (PTT) in Table~\ref{tab:ptt}. 
We compare the performance of the PTT with an alternative model where we replace the point-trajectory transformer with a multi-layer perceptron (MLP) of comparable parameter count. 
The results indicate a noteworthy reduction in AP and APH across all three categories.
Specifically, in the context of vehicle detection, there is a significant decrease of 2.61 in AP and 2.58 in APH.
This indicates the importance of the role played by our proposed point-trajectory transformer in effectively capturing and integrating sequential information for improved object detection accuracy.

\input{tab/abl_ppfeat}
\input{tab/abl_ptt}

\input{tab/abl_attention}
\smallskip{\noindent{\bf Effectiveness of the attention mechanisms in the proposed point-trajectory aggregator.}}
Table~\ref{tab:abl_attention} shows the importance of different attention designs in the point-trajectory aggregator that involves self-attention and cross-attention mechanisms.
The results show that simultaneous usage of both attention operations in the proposed point-trajectory aggregator achieves the best performance across all metrics.

\smallskip{\noindent{\bf Effectiveness of different lengths of the input frame.}}
We compare the performance of our method across different numbers of past frames, as presented in Table~\ref{tab:length}.
The results of mAPH on both L1 and L2 consistently show improvements as the number of past frames increases from 4 to 64. This emphasizes the importance of incorporating information from an extended time range and demonstrates the effectiveness of our method in integrating this valuable sequential data.
Furthermore, the performance shows a slight decline when utilizing 128 past frames, suggesting the inclusion of redundant information. Consequently, we choose 32 and 64 frames as a trade-off.

\input{tab/abl_length}

%% file: tab/abl_longshort.tex
\setlength{\tabcolsep}{0.05\linewidth}{
\begin{table}[t]
    \footnotesize
    \centering
    \begin{tabular}{cc|cccc}  
        \hline
        \multirow{2}{*}{Long} & \multirow{2}{*}{Short}  & L1 &L2 \\  
        & & mAP / mAPH & mAP / mAPH \\
        \hline\hline
        \xmark & \xmark &79.61/78.22 & 74.21/72.79\\
        \cmark & \xmark & 80.00/78.59 & 74.33/72.98 \\ 
        \xmark & \cmark & 79.96/78.55 & 74.77/72.93  \\  
        \cmark & \cmark & 81.38/80.04 & 75.88/74.59  \\
        \hline
    \end{tabular}
    \vspace{-1mm}
    \caption{\textbf{Ablation study on the effectiveness of the long-term and short-term memory}. The mean AP and APH are reported for L1 and L2 difficulties.
    }
    \label{tab:abl_longshort}
\end{table}
}

%% file: tab/abl_future.tex
\begin{table}[t]
    \centering
    \footnotesize
    \renewcommand{\arraystretch}{1.1}
    \setlength{\tabcolsep}{5pt}
    \begin{tabular}{l|ccccccc}
    \hline
      \multirow{2}{*}{Method}& \multicolumn{2}{c}{Vehicle} & \multicolumn{2}{c}{Pedestrian} & \multicolumn{2}{c}{Cyclist} \\
     & L1 & L2 & L1 & L2 & L1 & L2\\
    \hline\hline
    w/o Future &  81.31 & 74.00 & 81.80 & 74.73 & 75.98 & 73.82 \\ 
    w/ Future & 81.50 & 74.32 & 82.25 & 75.22 & 76.38 & 74.24  \\ 
    \hline
    \end{tabular}
    \vspace{-1mm}
    \caption{\textbf{Effectiveness of the proposed future encoding}. 
    APH on L1 and L2 difficulties for three categories are reported. 
    }
    \label{abl_future}
\end{table}

%% file: tab/abl_ppfeat.tex
\setlength{\tabcolsep}{0.03\linewidth}{
\begin{table}[t]
    \footnotesize
    \centering
    \begin{tabular}{cc|cc}  
        \hline
        Point-to- & Proposal-to- & L1 & L2\\
        Proposal  & Proposal  & mAP / mAPH & mAP / mAPH\\
        \hline\hline
        \xmark & \xmark & 79.99/78.58 & 74.34/72.98\\
        \cmark & \xmark & 80.13/78.71 & 74.46/73.10\\

        \xmark & \cmark & 81.15/79.79 &  75.58/74.27   \\  
        \cmark & \cmark & 81.38/80.04 &  75.88/74.59   \\
        \hline
    \end{tabular}
    \vspace{-1mm}
    \caption{\textbf{Effectiveness of the proposed Point-to-Proposal and Proposal-to-Proposal features}, which are defined in \eqref{eq:point} and \eqref{eq:proposal}, respectively. We report the mean AP and APH for the L1 and L2 difficulties.
    }
    \label{tab:abl_ppfeat}
\end{table}
}

%% file: tab/abl_ptt.tex
\setlength{\tabcolsep}{0.032\linewidth}{
\begin{table}[t]
    \footnotesize
    \centering
    \begin{tabular}{c|ccc}  
        \hline
        \multirow{2}{*}{Method}& Vehicle & Pedestrian & Cyclist\\
         & AP / APH & AP / APH & AP / APH\\
        \hline\hline
        MLP & 72.22/71.74 & 76.51/73.65 & 74.19/73.46 \\ 
        Transformer &74.83/74.32 & 77.93/75.22   & 74.89/74.24 \\ %
        \hline
    \end{tabular}
    \vspace{-1mm}
    \caption{\textbf{Analysis of different methods to encode the trajectory}. AP and APH scores of L2 difficulties for three categories are reported. 
    Transformer means our proposed point-trajectory transformer (PTT) as discussed in Section~\ref{sec:ptt}.}
    \label{tab:ptt}
\end{table}
}

%% file: tab/abl_attention.tex
\setlength{\tabcolsep}{0.045\linewidth}{
\begin{table}[t]
    \footnotesize
    \centering
    \begin{tabular}{cc|cccc}  
        \hline
\multirow{2}{*}{SA} & \multirow{2}{*}{CA}  & L1 &L2 \\  
        & & mAP / mAPH & mAP / mAPH \\  
        \hline\hline
        \xmark & \xmark & 80.00/78.58 & 74.31/72.95 \\ 
        \cmark & \xmark & 81.12/79.78 & 75.63/74.34 \\ 
        \xmark & \cmark & 81.15/79.79 & 75.58/74.27 \\
        \cmark & \cmark & 81.38/80.04 & 75.88/74.59 \\ 
        \hline
    \end{tabular}
    \vspace{-1mm}
    \caption{\textbf{Ablation study on different attention mechanisms in the proposed point-trajectory aggregator.} ``SA'' and ``CA'' denote the self-attention and cross-attention operations as shown in Figure~\ref{fig:pt_agg}, respectively.
    }
    \label{tab:abl_attention}
\end{table}
}

%% file: tab/abl_length.tex
\setlength{\tabcolsep}{0.04\linewidth}{
\begin{table}[t]
    \footnotesize
    \centering
    \begin{tabular}{c|cc} 
        \hline
        \multirow{2}{*}{Frame Length} & L1 & L2\\
         & mAP / mAPH & mAP / mAPH\\
        \hline\hline
        4 & 80.47/79.06 & 74.78/73.43\\
        8 & 80.56/79.25 & 74.95/73.66  \\
        16 & 81.13/79.80 & 75.58/74.29 \\  
        32 & 81.38/80.04 & 75.88/74.59 \\
        64 & 81.54/80.21 & 76.05/74.75   \\
        128 & 81.46/80.11 & 75.95/74.68  \\
        \hline
    \end{tabular}
    \vspace{-1mm}
    \caption{\textbf{Comparisons for different lengths of input frames.} We report the mean APH scores among three categories for the level 1 and level 2 difficulties.
    }
    \label{tab:length}
\end{table}
}

%% file: sec/4_Conclusion.tex
\section{Conclusions}
In this paper, we present a Point-Trajectory Transformer (PTT) for efficient temporal 3D object detection. We find that 1) leveraging multi-frame point clouds can lead to memory overhead, and 2) considering multi-frame proposal trajectories can be efficient and effective, where these two observations are not widely studied in prior works.
To this end, our PTT efficiently establishes connections between single-frame point clouds and multi-frame proposals, facilitating the utilization of rich LiDAR data with reduced memory overhead. By exploiting fewer points for learning effective representations, our approach allows processing information from more frames without encountering memory issues.
Meanwhile, we propose long-term, short-term, and future-aware encoders to enhance feature learning over temporal information and enable the ability to generate future-aware features via future trajectories, followed by the proposed point-trajectory aggregator to integrate point clouds and proposals effectively.
Our method performs favorably against state-of-the-art approaches on the challenging Waymo Open Dataset, using more frames but smaller memory overhear and faster runtime, showcasing the effectiveness of the proposed approach.